\documentclass{article} 
\usepackage{iclr2025_conference,times}

\usepackage{amsmath,amsfonts,bm}

\def\eqref#1{equation~\ref{#1}}

\def\1{\bm{1}}

\DeclareMathAlphabet{\mathsfit}{\encodingdefault}{\sfdefault}{m}{sl}
\SetMathAlphabet{\mathsfit}{bold}{\encodingdefault}{\sfdefault}{bx}{n}

\usepackage{hyperref}
\usepackage{url}
\iclrfinalcopy
\usepackage{times}
\usepackage{microtype}
\usepackage[dvipsnames]{xcolor} 
\usepackage[most]{tcolorbox}
\usepackage{xcolor}
\usepackage{natbib}

\definecolor{lightblue}{HTML}{B5E4F5}
\definecolor{darkblue}{HTML}{79BAD1}

\definecolor{lightpurple}{HTML}{EAE1ED}

\definecolor{purple}{HTML}{C9B4CF}

\usepackage{float}
\usepackage{enumitem}
\usepackage{hyperref}
\usepackage{comment}
\usepackage{graphicx}
\usepackage{xcolor}
\usepackage{booktabs}
\usepackage{amsmath,amssymb}
\usepackage{abstract}
\usepackage{balance}
\usepackage{tabularx,ragged2e}

\title{Can Large Language Models Improve Accuracy on Mathematical Tasks Using Flawed \\ Thinking?}

\author{Saraswathy Amjith, Mihika Dusad, Neha Muramalla, Shweta Shah  \\
MIT \\
\thanks{Code available at \url{https://github.com/saraswathyamjith/FlawedMathReasoningRL}}
\texttt{\{swathy, mdusad, neham, sshah27\}@mit.edu}} 

\begin{document}

\maketitle

\begin{abstract}
Chain-of-thought (CoT) prompting has become central to mathematical reasoning in large language models, yet models remain brittle to early errors: a single arithmetic slip or unjustified inference typically propagates uncorrected to an incorrect final answer. We investigate whether training on intentionally flawed reasoning traces can teach models to detect and recover from such errors without degrading standard problem-solving ability. Using competition-level problems from MATH-lighteval, we generate CoT prefixes containing exactly one controlled error, either a calculation error (sign flips, dropped terms) or a reasoning error (misapplied rules, unjustified logical steps), and fine-tune Qwen3-4B with GRPO using a binary final-answer reward. Our Mixed-CoT-RL model matches standard RL on clean problems (41\% vs.\ 41\%) while substantially outperforming it on problems prefilled with flawed reasoning (24\% vs.\ 19\%). Notably, clean-only RL fine-tuning \emph{degrades} robustness below the untuned baseline (19\% vs.\ 20\%), indicating that conventional training increases susceptibility to misleading prefills. Among error types, training on reasoning errors yields greater robustness gains than calculation errors alone, with mixed training performing best. These findings demonstrate that exposure to flawed traces during training can improve error-recovery behavior without sacrificing accuracy, suggesting a path toward more robust mathematical reasoning in LLMs.

\end{abstract}

\section{Introduction}
Recently, we have seen large language models (LLMs) achieve incredibly strong performances across a wide range of reasoning tasks. Strong performance of models is typically measured by quantitative metrics such as accuracy on a dataset, and typical reasoning tasks require logical thinking, complex analysis, and multi-step problem-solving. The most common benchmarks for reasoning tasks are math and coding-based tasks. Especially since ChatGPT was released in November 2022, LLMs have improved rapidly at solving difficult math and coding problems, and much of this progress can be attributed to post-training methods, which seek to fine-tune a base model after it has already been trained by essentially highlighting specific example responses it should focus on emulating. Well-known examples of post-training processes include Reinforcement Learning from Human Feedback (RLHF) \citep{christiano2017deep}, which works by having humans rate responses generated by the model, and Direct Preference Optimization (DPO) \citep{rafailov2023direct}, which aligns model outputs with human preferences using a simpler objective function and eliminating the need for a separate reward model.  

However, a major limitation of existing post-training methods is that they primarily optimize LLMs to produce a correct final answer rather than the correct intermediate chain-of-thought (CoT) used to derive it. As a result, even reasoning models can be brittle: if an early error is introduced, for example, an arithmetic slip or an unjustified logical step, the model typically propagates the mistake rather than correcting it. This is especially problematic in mathematics, where a single local error can invalidate an entire derivation. Ensuring that LLMs can identify and recover from their own mistakes is vital for users who rely on step-by-step solutions for both correctness and understanding.  

\begin{figure}[H]
    \centering
    \includegraphics[scale = 0.5]{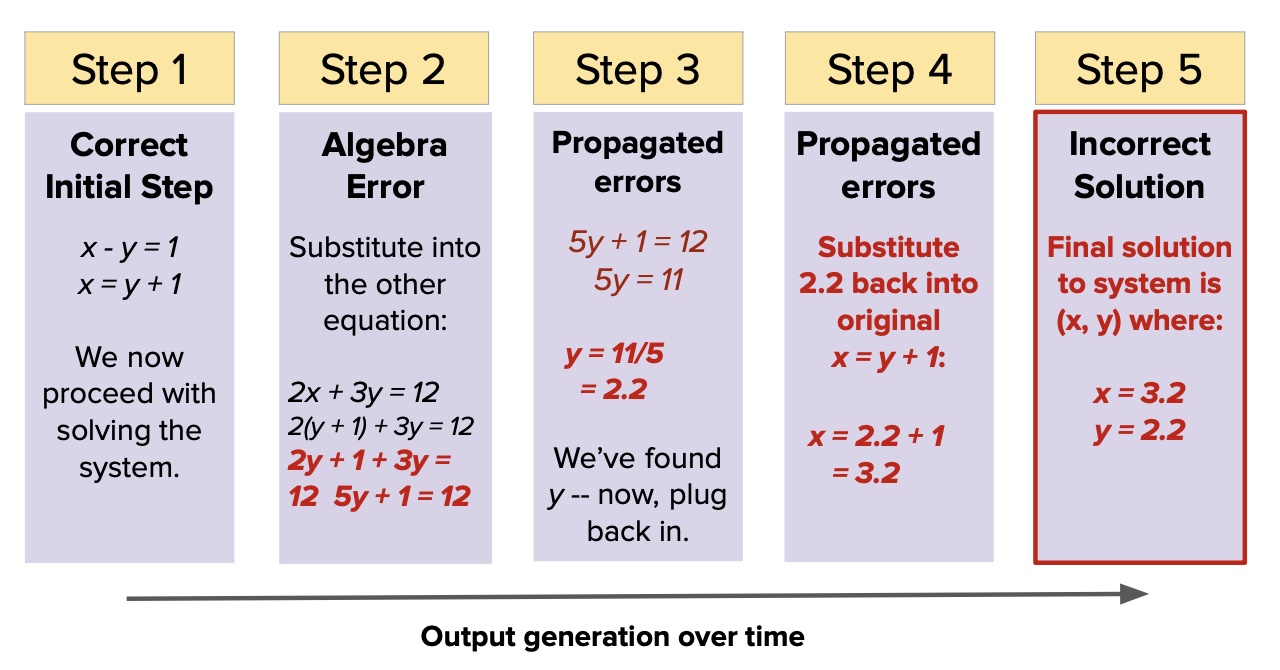} 
    \caption{Example of error propagation in CoTs.}
    \label{fig:finetune}
\end{figure}

Reliable step-by-step reasoning in LLMs has important benefits beyond simply producing correct answers. For mathematics researchers, LLMs can assist by verifying intermediate calculations, exploring alternative solution strategies, and creating derivations more quickly than manual methods. For learners, LLMs can function as interactive study tools: students can follow detailed problem-solving steps, receive immediate explanations for each step, and gain intuitive understanding of complex concepts. By improving the reasoning process, LLMs can improve mathematical comprehension and support more effective learning and research workflows.

 In this paper, we introduce a simple classification of flawed reasoning types and study how different error families affect a model’s ability to recover from misleading CoTs. Specifically, for each math problem, we prefill the Chain-of-Thought with one of two error types (\emph{calculation errors} or \emph{reasoning errors}) and evaluate how the model generates a corrected reasoning sequence and final answer.

\textbf{Calculation errors} include sign flips, dropped terms, or incorrect simplifications.
\textbf{Reasoning errors} include unjustified inferences, misapplied rules, or broken invariants.
We then train models under mixtures of clean and flawed trajectories and evaluate how exposure to each error type influences robustness to adversarial or misleading prefills.

Our primary contributions are:

\textbf{(1)} We formulate a process-level training setup for mathematical reasoning that explicitly injects flawed CoTs and encourages the model to recover from them.

\textbf{(2)} We explore two major families of errors—calculation errors and reasoning errors—and measure how training on each type affects robustness.

\textbf{(3)} We outline a reinforcement learning pipeline based on a binary reward model and GRPO that can be run on modest hardware while still enabling fine-grained process-level supervision.

\section{Background}

Post-training methods such as RLHF \citep{christiano2017deep} and DPO \citep{rafailov2023direct} have significantly improved LLM performance across reasoning tasks. These approaches optimize models to produce preferred outputs but typically supervise only the final answer, leaving intermediate reasoning steps unsupervised. As a result, LLMs can propagate early mistakes in multi-step problems, particularly in mathematics, where local errors can invalidate entire derivations \citep{uesato2022solving}. Recent efforts, including \emph{Outcome Reward Models (ORMs)} and \emph{Process Reward Models (PRMs)}, attempt to score reasoning steps sequentially, showing that step-level supervision improves faithfulness and interpretability \citep{lightman2024lets, wang2024prm}. However, these methods generally rely on clean reasoning traces and do not explore how exposing models to flawed or misleading steps affects learning.

To counter this, Peng et al. \citep{peng2025recap} introduce the RECAP framework, which strengthens models' ability to withstand missteps in reasoning. RECAP trains models to recover from \emph{flawed} reasoning prefixes: short, intentionally incorrect CoTs that precede correct demonstrations. In safety-alignment settings, this process-level supervision reduces vulnerability to jailbreaks, refusal attacks, and adversarial prompts, outperforming standard RLHF/DPO-style post-training. However, RECAP treats all flawed traces uniformly and focuses on safety and moral reasoning, where failures are discrete and easy to label. Errors are binary (“violates safety” vs. “does not”) and failures are localized and categorical (e.g., providing harmful instructions, endorsing discrimination). This is fundamentally different from mathematical or scientific reasoning, where multiple distinct trajectories can lead to correct or incorrect outcomes, and it is unclear which kinds of flawed traces are most useful for training. In fact, the authors explicitly highlight mathematical reasoning as an open challenge.

Thus, in this paper, we take on the challenge of applying the RECAP idea to \emph{mathematical} reasoning. Our work extends this flawed-prefix framework by separating injected errors into \emph{calculation} and \emph{reasoning} types and empirically studying how training on each affects robustness. This focus connects to a broader line of work exploring how models detect, correct, or learn from mistakes in mathematical reasoning. Several other studies have also been done in this domain, which explores self-correction mechanisms for mathematical reasoning. For example, \emph{S$^3$c-Math} encourages models to spontaneously detect and correct step-level errors, improving overall correctness on arithmetic and algebra problems \citep{ssc_math2024}. Similarly, \emph{LEMA} uses incorrect reasoning traces paired with corrected solutions for fine-tuning, mimicking human learning from mistakes \citep{lema2023}. Stepwise verification methods such as \emph{StepCo} and pedagogically inspired frameworks like \emph{PedCoT} reinforce the importance of multi-step reasoning robustness by enabling models to evaluate and correct intermediate steps \citep{xu2024pedcot}. These works also motivate our approach of explicitly classifying and training on different types of flawed reasoning.

Looking beyond the methods stated above, mathematical reasoning in machine learning has also been shaped by new, increasingly complex datasets. The emergence of datasets such as \emph{RMath} \citep{hu2025rmath}, \emph{MuMath} \citep{you2024mumath}, and \emph{HighMATH} \citep{liu2025highmath} reflects the community’s growing interest in complex, multi-step mathematical reasoning beyond arithmetic. These datasets highlight that reasoning involves not only computation but also logical deduction, planning, and symbolic manipulation. However, none of these datasets evaluate whether models can recover from misleading or subtly incorrect problem prompts, a gap we address by introducing a perturbed-prompt benchmark that tests robustness under intentionally corrupted math questions.

While orthogonal to our method, another line of work enhances mathematical reasoning through hybrid symbolic and neuro-symbolic frameworks, which provide alternative methods for enhancing mathematical reasoning. Approaches like \emph{SymbCoT}, \emph{CoMAT}, and \emph{PoT} integrate symbolic representations or programmatic computation with LLM reasoning, improving logical fidelity and reducing arithmetic errors \citep{xu2024symbcot, leang2024comat, chen2022pot}. Similarly, systems such as \emph{LogicLM} and \emph{Lean-STaR} translate natural-language problems into symbolic form, iteratively refining solutions with solver verification \citep{panLogicLM23, zhou2023leanstar}. These works show that alternative paradigms can enhance correctness, but they do not systematically study how training on flawed traces affects process-level reasoning, which is the focus of our study.

Beyond LLM-specific methods, research in cognitive science and curriculum learning supports the idea that controlled exposure to errors can enhance learning. Introducing errors in a structured manner encourages exploration, critical evaluation, and robustness, principles that motivate our systematic study of flawed reasoning types in training LLMs.

\section{Methods}

\subsection{Problem Setup}

Let \text{LLM} denote the policy model that generates reasoning trajectories. Given a query $q$, the model produces a multi-step chain-of-thought
\[
\mathbf{x} = (x_1, x_2, \ldots, x_T),
\]
where $x_t$ is the $t$-th reasoning step and $T$ is the trajectory length. We use $\mathbf{x}_{1:t}$ to denote the prefix up to step $t$.

Given a dataset of math queries $\mathcal{Q} = \{ q_1, \ldots, q_N \}$, we construct \emph{flawed} trajectories $\tilde{\mathbf{x}}^{(i)}$ by injecting exactly one controlled error at the first step step. Errors are categorized as either calculation errors or reasoning errors.
Calculation error examples include dropped terms, sign flips, or incorrect simplifications. Reasoning error examples include  misapplied rules, unjustified inferences, or logic jumps. 
A mixture hyperparameter $\alpha \in [0,1]$ controls the fraction of training examples that contain flawed prefixes (reasoning or calculation):
\[\mathcal{D}_{\text{train}} = 
\{ (q_i, \mathbf{x}^{(i)}_{\text{train}}) \mid i = 1, \ldots, N \},\]
where each $\mathbf{x}^{(i)}_{\text{train}}$ is clean with probability $1-\alpha$ and flawed with probability $\alpha$. This mixture allows the model to learn from both correct reasoning and examples containing subtle errors.
\begin{figure}[H]
    \centering
    \includegraphics[scale = 0.56]{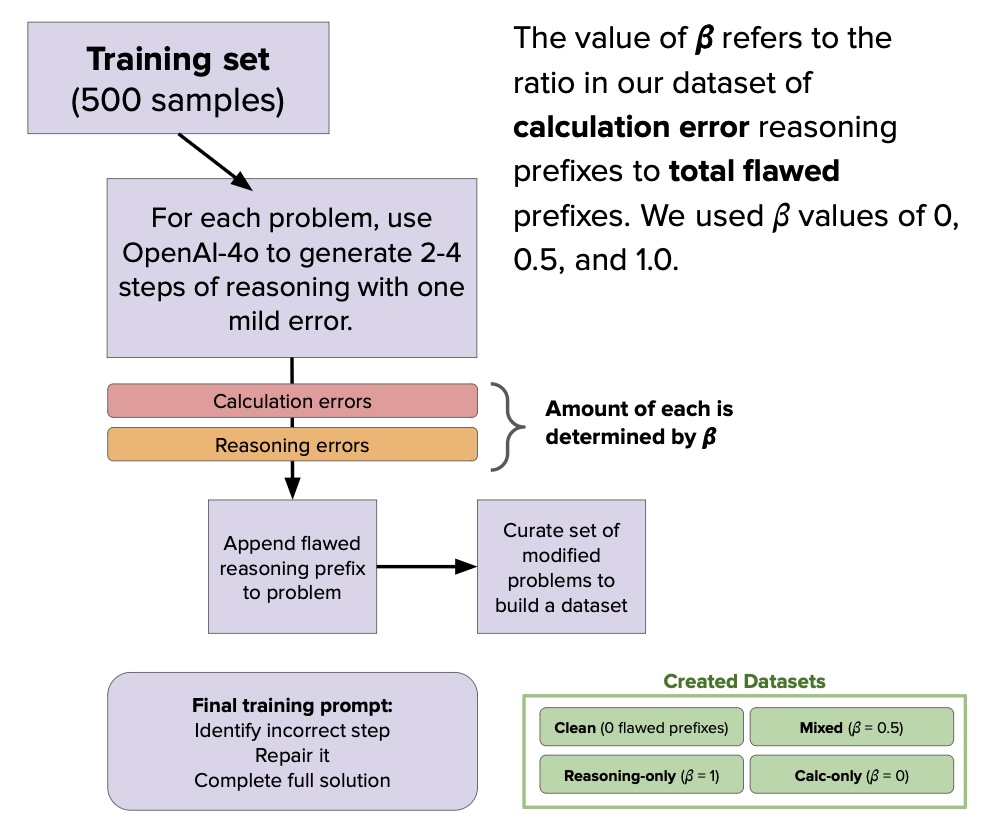} 
    \caption{High-level training flow with flawed prefix injection and process rewards.}
    \label{fig:finetune}
\end{figure}

During fine-tuning, the model parameters $\theta$ are optimized using a \textbf{binary final-answer reward}:
\[
R(q, \mathbf{x}) =
\begin{cases} 
1 & \text{if the final answer is exactly correct},\\
-1 & \text{otherwise}.
\end{cases}
\]

The training objective is thus
\[
\max_\theta \ \mathbb{E}_{q \sim \mathcal{Q}} \ \mathbb{E}_{\mathbf{x} \sim \text{LLM}_\theta(q)} \left[ R(q, \mathbf{x}) \right],
\]
where the presence of flawed trajectories encourages the model to learn corrective behavior following corrupted prefixes $\tilde{\mathbf{x}}_{1:t}$.

\begin{figure}[h]
    \centering
    \includegraphics[scale = 0.46]{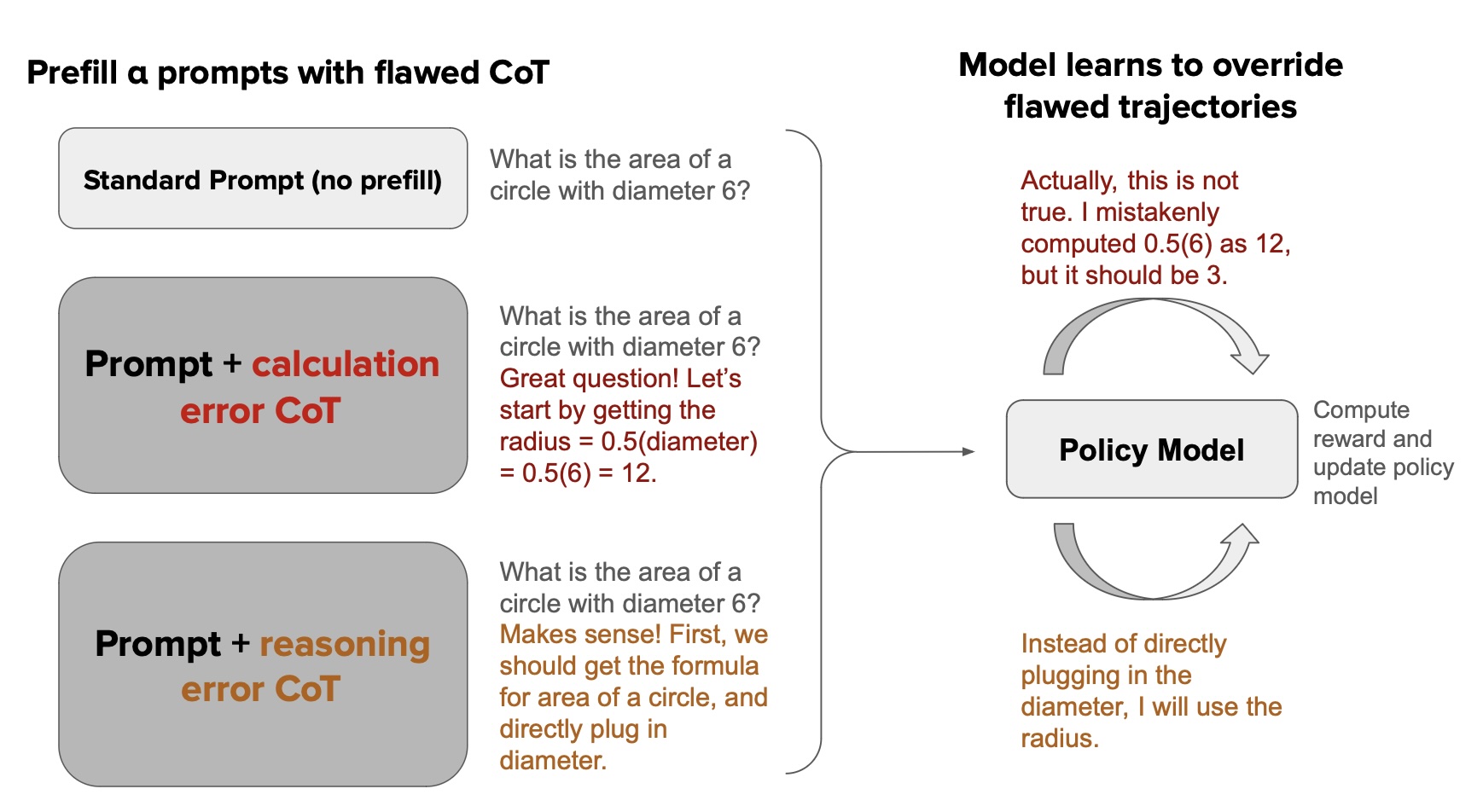} 
    \caption{High-level training flow with flawed prefix injection and process rewards.}
    \label{fig:finetune}
\end{figure}%
\begin{tcolorbox}[
  colback=lightblue,
  colframe=darkblue,
  title=Flawed Solution Prompt Template,
  fonttitle=\bfseries,
  arc=2mm,
  boxrule=0.5pt,
  left=4pt, right=4pt,
  top=0pt, bottom=0pt
]
\textbf{Task.} Generate a \emph{FLAWED} mathematical solution for this problem.

\medskip

\textbf{Problem.} \emph{[problem]}

\medskip

\textbf{Correct Answer (do NOT arrive at this).} \emph{[correct\_answer]}

\medskip

\textbf{Requirements.}
\begin{enumerate}
  \item Make \textbf{ONE} subtle error (reasoning error, calculation mistake, sign error, or wrong formula).
  \item Continue reasoning logically from that error.
  \item Arrive at a \textbf{WRONG} answer.
  \item Keep it concise (only a single step).
  \item Do \textbf{not} mention there is an error.
  \item Do \textbf{not} include \verb|\boxed{}| — leave the solution incomplete.
\end{enumerate}

Generate the flawed reasoning (no final boxed answer).
\end{tcolorbox}

\begin{tcolorbox}[
  colback=lightpurple,
  colframe=purple,
  title=Qwen-3-4b RL Prompt Template,
  fonttitle=\bfseries,
  arc=2mm,
  boxrule=0.5pt,
  left=4pt, right=4pt,
  top=0pt, bottom=0pt
]
\textbf{System.} You are a helpful math tutor. Solve problems step by step.  
If you notice errors in your reasoning, correct them.  
Put your final answer in \(\boxed{\text{answer}}\) format.

\medskip

\textbf{User.} Solve the following problem:  
\emph{[problem]}

\medskip

\textbf{Assistant (prefilled chain-of-thought).}  
\emph{[flawed\_cot]}
\end{tcolorbox}

\subsection{Models and Training}

We use open-source language models that have not been instruction tuned. Our primary model is \emph{Qwen3-4B} \citep{yang2025qwen3technicalreport} to verify that observed trends are not architecture-specific.

We fine-tune the backbone models using reinforcement learning with GRPO, a policy-gradient method that maximizes expected reward while being more resource-efficient than classical PPO-style RL. Our GRPO training run utilized the Qwen/Qwen3-4B base model. It applied Low-Rank Adaptation (LoRA) to the base model with the following configuration:
\[
r=128,\hspace{6pt} \text{lora}_{\alpha}=32, \text{lora}_{\text{dropout}}=0.05 
\]
The training algorithm was built on a binary final-answer reward system.
We configured GRPO with a per-device batch size of 2 and gradient accumulation steps of 2, yielding an effective batch size of 4. We set the number of generations per prompt to 2, allowing the model to sample multiple candidate completions for each training example. The learning rate was set to 1e-5, and training proceeded for 150 steps. We used a temperature of 0.7 and top-p sampling at 0.9 to encourage diverse yet coherent generations. The KL penalty coefficient (beta) was set to 0.0, allowing the model to deviate freely from the reference policy during optimization, a choice motivated by our goal of teaching substantively new error-recovery behavior rather than staying close to the base model's distribution. Maximum completion and prompt lengths were set to 1024 and 2048 tokens, respectively, to accommodate the multi-step reasoning traces typical of competition mathematics problems.

\noindent\textbf{Reward design:}  
We use a \emph{binary final-answer reward}: \[
r =
\begin{cases}
1, & \text{if the final answer is exactly correct},\\
-1, & \text{otherwise}.
\end{cases}
\]
A LLM-as-a-judge model was initially considered but later discarded upon producing subpar results. Initially, we had two training runs comparing a LLM-as-a-judge model versus the binary reward and the accuracies were 31\% and 34\%, respectively. The LLM-as-a-judge model was modeled as a partial-reward-model to give partial rewards for preliminary steps, but we found that a binary reward yielded better results. Thus, learning is driven solely by the correctness of the final answer

\subsection{Verification and Evaluation}
During training we record the model's intermediate reasoning for logging and analysis, but these traces are not used for computing reward. Learning is driven solely by the correctness of the final answer.
Candidate solutions are verified using numeric checks, by sampling values consistent with the problem constraints and comparing the model's output to a ground-truth evaluator.
The model is required to mark its final answer in a canonical format (\textbackslash boxed\{...\}) to enable reliable extraction and scoring.
\subsubsection{Benchmarks}
We train on the \textbf{MATH-lighteval} \citep{hendrycksmath2021} benchmark of high-school level competition math problems, selecting a randomized training set of 500 samples. \textbf{MATH-lighteval} consists of problems from mathematics competitions, including the AMC 10, AMC 12, AIME, and more. This dataset consists of 12,500 examples of diverse, multi-step symbolic reasoning problems that stress long-horizon mathematical reasoning. 
\begin{tcolorbox}[
    colback=gray!10,
    colframe=gray!50,
    title=\textbf{Example Problem},
    fonttitle=\bfseries,
    boxrule=0.5pt,
    arc=2mm
]
Let $P_1$ be a polygon in the coordinate plane with vertices
\[
(x_1,y_1), (x_2,y_2), \dots, (x_{33},y_{33}),
\]
listed in order. Suppose that the sum of the $x$-coordinates of the vertices of $P_1$ is
\[
x_1 + x_2 + \cdots + x_{33} = 99.
\]
Construct polygon $P_2$ whose vertices are the midpoints of consecutive vertices of $P_1$, and similarly construct polygon $P_3$ whose vertices are the midpoints of consecutive vertices of $P_2$. What is the sum of the $x$-coordinates of the vertices of $P_3$?
\end{tcolorbox}

We create two disjoint held-out splits derived from MATH-lighteval: one for development and one for final reporting. Test prompts are deterministic templates with canonical answer checking. Unless otherwise specified, we evaluate with temperature $T = 0.6$ and use $n$-sample decoding when exploring sampling-based improvements. Note, that Math-500 is a subset of 500 problems from MATH-lighteval, which we hold out as our test set. 

We evaluate five different models: \textbf{Pretrained (PT)}, the untuned base Qwen3-4B model; \textbf{Ablation-RL}, GRPO trained only on the same problems, without the flawed CoT, with a binary final-answer reward; \textbf{Flawed-CoT RL (Mixed)}, GRPO with binary rewards trained on mixtures of clean and flawed trajectories as described above; \textbf{Flawed-CoT RL (Calc-only)}, identical to the Mixed setting but restricted to calculation-error prefills only; and \textbf{Flawed-CoT RL (Reasoning-only)}, identical to the Mixed setting but restricted to reasoning-error prefills only.

\subsection{Perturbed-Math-100 Dataset Creation}
To evaluate robustness to misleading reasoning, we constructed the Perturbed-Math-100 dataset by generating flawed chain-of-thought prefixes for 100 held-out problems from MATH-500. For each problem, we prompted GPT-4o-mini to produce a single reasoning step containing exactly one subtle error—either a calculation error (e.g., sign flip, dropped term) or a reasoning error (e.g., misapplied theorem, unjustified inference), while continuing logically from that mistake without revealing the error's presence. These flawed prefixes were then prepended to the model's context as an assistant prefill, simulating scenarios where the model must continue from corrupted intermediate reasoning. This dataset enables direct measurement of error-recovery capability: models that blindly propagate the injected error will arrive at incorrect answers, while models with robust self-correction will detect the inconsistency and recover toward the correct solution.
\begin{tcolorbox}[
    colback=gray!10,
    colframe=gray!50,
    title=\textbf{perturbed-math-100 example},
    fonttitle=\bfseries,
    boxrule=0.5pt,
    arc=2mm
]
A square and a regular heptagon are coplanar and share a common side $\overline{AD}$, as shown. 
What is the degree measure of angle $\angle BAC$? Express your answer as a common fraction.
\begin{center}
\end{center}
\medskip
\noindent\textbf{Start of a solution.}
The measure of each interior angle in a regular $n$-gon is 
\[
\frac{180n}{(n-2)} \text{ degrees}.
\]
\end{tcolorbox}
\section{Results}
We first report standard MATH-500 accuracy on the first 100 problems subset for the baseline model, RL without flawed CoT ablation, Calculation Errors Flawed Cot RL , Reasoning Errors Flawed CoT RL, and Mixed Errors Flawed CoT Rl with both Calculations and Reasoning Errors. All accuracy measurements are done on a held-out dataset of Math-500. 

Our primary ablation compares Calculation Errors Flawed Cot RL, Reasoning Errors Flawed CoT RL, and Mixed Errors Flawed CoT Rl to RL without flawed CoT ablation to isolate the causal effect of training on flawed reasoning prefixes. Both conditions use identical binary final-answer rewards and identical GRPO training hyperparameters; the only difference is whether the training data contains flawed-COT errors. If improvements were driven solely by “more RL on math,” the Ablation-RL should match or exceed Flawed-CoT RL. A consistent advantage otherwise would indicate that exposure to flawed prefixes improves error-recovery reasoning.

We report accuracy as the fraction of correctly solved problems out of 100 held-out examples. Since each problem is an independent Bernoulli trial with success probability $p$, we model accuracy as a binomial proportion and compute standard errors as $
\text{SE} = \sqrt{\frac{\hat{p}(1-\hat{p})}{n}}$,
where $\hat{p}$ is the observed accuracy and $n = 100$. All error bars in our figures represent $\pm 1$ standard error under this assumption.

\subsection{RL Training Rewards Over Time on various models}
\begin{figure}[H]
    \centering
    \includegraphics[scale = 0.45]{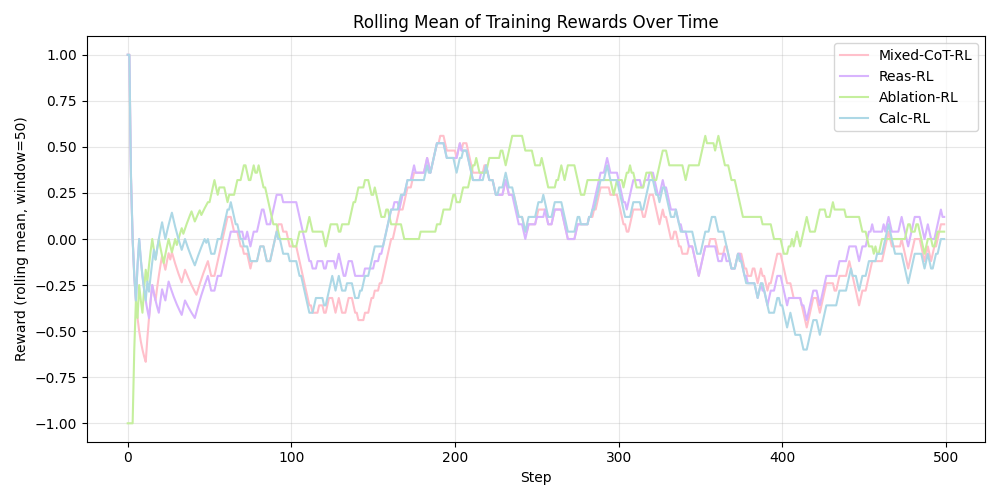} 
    \caption{Rolling Mean of Training Rewards Over Time During GRPO RL}
    \label{fig:finetune}
\end{figure}

\subsection{Overall Accuracy on MATH-500}

\begin{figure}[H]
    \centering
    \includegraphics[scale = 0.65]{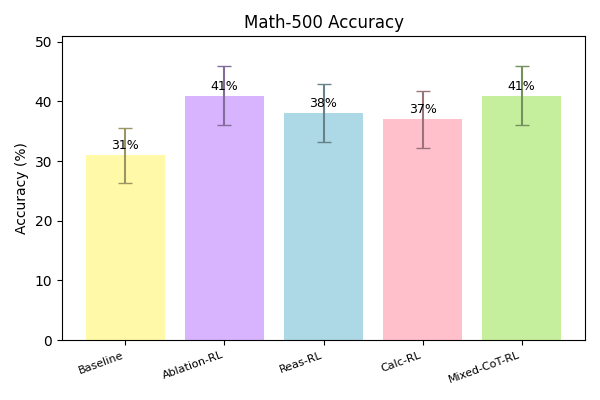} 
    \caption{Math-500 Accuracy. Error bars show $\pm 1$ standard error under binomial sampling ($\text{SE} = \sqrt{\hat{p}(1-\hat{p})/100}$).}
    \label{fig:finetune}
\end{figure}

\subsection{Robustness to Perturbations}
We created a dataset with flawed reasoning following the prompt to assess whether the models will continue in this incorrect line of reasoning or self-correct. We call this Perturbed-Math-100, and report accuracy on the baseline model, RL without flawed CoT ablation, Calculation Errors Flawed Cot RL , Reasoning Errors Flawed CoT RL, and Mixed Errors Flawed CoT Rl with both Calculations and Reasoning Errors. 

\begin{figure}[H]
    \centering
    \includegraphics[scale = 0.65]{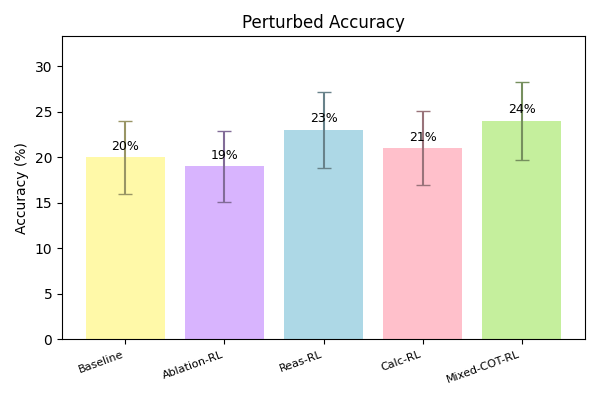} 
    \caption{Peturbation-Math-100 Accuracy Accuracy. Error bars show $\pm 1$ standard error under binomial sampling ($\text{SE} = \sqrt{\hat{p}(1-\hat{p})/100}$).}
    \label{fig:finetune}
\end{figure}

\subsection{Effect of Error Type}
Using the error-type labels from our data construction, we  compare training mixtures dominated by calculation errors, reasoning errors, or a balanced mix of both.

\section{Discussion}

Our experiments reveal a key result: training on intentionally flawed chain-of-thought trajectories improves robustness to misleading CoT prefills \emph{without sacrificing standard accuracy}. On the MATH-500 benchmark, Mixed-COT-RL achieved 41\% accuracy—matching the Ablation RL ablation exactly, while on the Perturbed-Math-100 benchmark, Mixed-COT-RL achieved 24\% compared to only 19\% for Ablation RL. This 5-percentage point improvement in robustness, which comes at no cost to standard performance, suggests that Flawed-CoT training teaches systematic error-recovery behavior that complements rather than competes with general mathematical reasoning ability.

One might expect that training on corrupted reasoning traces would degrade the model's ability to solve normal problems, either by introducing noise into learned solution strategies or by teaching the model to ``expect'' errors where none exist. Instead, we observe that the model successfully does both: it learns to detect and correct errors when they are present while maintaining its baseline problem-solving capabilities on unperturbed inputs. 

 Ablation RL, despite achieving the highest standard accuracy (41\%), performed \emph{worse} than the untuned baseline on perturbed problems (19\% vs.\ 20\%). This result suggests that standard RL fine-tuning may inadvertently increase susceptibility to misleading prefills. We hypothesize that optimizing purely for final-answer correctness on clean trajectories reinforces the model's tendency to continue confidently from any given prefix, rather than critically evaluating whether that prefix contains errors.

In contrast, all three Flawed-CoT RL variants outperformed both baseline and  Ablation-RL on perturbed problems, confirming that the robustness improvements stem specifically from exposure to flawed traces. 

On standard MATH-500 accuracy, we observe that Ablation-RL (41\%) and Mixed-COT-RL (41\%) outperform both Reasoning-Errors-RL (38\%) and Calculation-Errors-RL (37\%). This ordering suggests that training exclusively on one error type may slightly impair standard problem-solving, perhaps because the model overfits to detecting that specific error family. The mixed training regime appears to strike the best balance, matching clean RL on standard problems while exceeding it on perturbed ones.

The 10-percentage-point improvement over baseline (31\% $\to$ 41\%) achieved by both Ablation RL and Mixed-COT-RL confirms that GRPO fine-tuning with binary rewards substantially improves mathematical reasoning. Critically, Flawed-CoT training captures these gains while additionally conferring robustness benefits.

On perturbed problems, Reasoning-Errors-RL (23\%) outperformed Calculation-Errors-RL (21\%), with Mixed-COT-RL (24\%) achieving the best overall robustness. This pattern suggests that training on reasoning errors, such as misapplied rules or unjustified logical jumps, may better prepare models to handle diverse perturbations than training on arithmetic mistakes alone. Reasoning errors require the model to evaluate logical validity rather than simply recompute numerical results, potentially fostering more general critical-evaluation skills.

The standard errors for the perturbed accuracy results are large due to the limited sample size ($n=100$). While all RL-trained models achieve statistically significant improvements over the baseline (31\%), the error bars for Flawed-CoT RL variants overlap substantially with Ablation-RL, preventing strong conclusions about flawed-CoT effects. Future work with larger evaluation sets would be needed to establish whether the robustness advantage of flawed-CoT training over standard RL fine-tuning is statistically significant.

The combination of maintained standard accuracy and improved robustness has significant practical implications.
Users frequently provide partial or incorrect solutions and ask models to continue or debug them. Students learning mathematics may supply work containing errors and expect the model to identify mistakes rather than propagate them. Our results indicate that Flawed-CoT RL produces models better suited to this role, as they are more likely to recognize and correct upstream errors rather than blindly extending flawed reasoning. As LLMs are deployed in agentic pipelines where they iteratively refine their own outputs, the ability to recover from self-generated errors becomes critical. Models trained on flawed trajectories may be better equipped to catch and correct their own mistakes during multi-turn reasoning, reducing error accumulation across extended problem-solving sessions.

\subsection{Limitations}

In drawing our conclusions, we acknowledge certain limitations. Firstly, our evaluation is limited to only 100 problems for both standard and perturbed accuracy, due to compute constraints. Furthermore, the confidence intervals—particularly for perturbed results—are relatively wide. Conducting larger-scale experiments could potentially strengthen the statistical confidence in our findings.

Another limitation of our work is that the flawed prefixes are generated by GPT-4o-mini and may not fully capture the natural distribution of problem-solving errors that arise during model generation or from humans doing tasks. The single-error constraint also limits ecological validity, as real-world mistakes run the risk of compounding across multiple steps.

Finally, we evaluate our models only on one use case, high-school competition mathematics, and we test on a single model architecture (Qwen3-4B). How our results generalize to other mathematical domains, reasoning tasks, and model types and scales remains to be established.

\subsection{Future Work}

Several directions emerge from this work. First, scaling our experiments to larger models and datasets could help establish whether the benefits of Flawed-CoT training persist, and possibly improve, at scale. Learning strategies that progressively increase the difficulty of the error or introduce multiple errors per trajectory may further improve the robustness. Extending this framework to other reasoning domains, including code generation, scientific problem-solving, and formal theorem proving, could also be used to assess the generality of our findings.

\subsection{Conclusion}
We have demonstrated that training language models on intentionally flawed mathematical reasoning traces substantially improves their ability to recover from misleading CoT prefills without degrading standard problem-solving performance. Our Mixed-COT-RL run matched Ablation RL in terms of MATH-500 accuracy (41\%), while substantially outperforming it on perturbed problems (24\% vs.\ 19\%). This improvement does not arise from standard RL fine-tuning on clean data alone—indeed, ablation RL slightly \emph{increases} susceptibility to perturbations, suggesting that explicit exposure to errors teaches systematic corrective behavior. These findings motivate further research into flawed-CoT training for reasoning in various domains.

\newpage
\vspace{0.5em}

\bibliography{references}

@misc{yang2025qwen3technicalreport,
      title={Qwen3 Technical Report}, 
      author={An Yang and Anfeng Li and Baosong Yang and Beichen Zhang and Binyuan Hui and Bo Zheng and Bowen Yu and Chang Gao and Chengen Huang and Chenxu Lv and Chujie Zheng and Dayiheng Liu and Fan Zhou and Fei Huang and Feng Hu and Hao Ge and Haoran Wei and Huan Lin and Jialong Tang and Jian Yang and Jianhong Tu and Jianwei Zhang and Jianxin Yang and Jiaxi Yang and Jing Zhou and Jingren Zhou and Junyang Lin and Kai Dang and Keqin Bao and Kexin Yang and Le Yu and Lianghao Deng and Mei Li and Mingfeng Xue and Mingze Li and Pei Zhang and Peng Wang and Qin Zhu and Rui Men and Ruize Gao and Shixuan Liu and Shuang Luo and Tianhao Li and Tianyi Tang and Wenbiao Yin and Xingzhang Ren and Xinyu Wang and Xinyu Zhang and Xuancheng Ren and Yang Fan and Yang Su and Yichang Zhang and Yinger Zhang and Yu Wan and Yuqiong Liu and Zekun Wang and Zeyu Cui and Zhenru Zhang and Zhipeng Zhou and Zihan Qiu},
      year={2025},
      eprint={2505.09388},
      archivePrefix={arXiv},
      primaryClass={cs.CL},
      url={https://arxiv.org/abs/2505.09388}, 
}

@inproceedings{christiano2017deep,
  title     = {Deep Reinforcement Learning from Human Preferences},
  author    = {Christiano, Paul F. and Leike, Jan and Brown, Tom and Martic, Miljan and Legg, Shane and Amodei, Dario},
  booktitle = {Advances in Neural Information Processing Systems},
  year      = {2017},
  url       = {https://arxiv.org/abs/1706.03741}
}

@inproceedings{rafailov2023direct,
  title     = {Direct Preference Optimization: Your Language Model is Secretly a Reward Model},
  author    = {Rafailov, Rafid and Sharma, Abhishek and Mitchell, Eric and Ermon, Stefano and Manning, Christopher D. and Finn, Chelsea},
  booktitle = {Advances in Neural Information Processing Systems},
  year      = {2023},
  url       = {https://arxiv.org/abs/2305.18290}
}

@article{uesato2022solving,
  title   = {Solving Math Word Problems with Process- and Outcome-Based Feedback},
  author  = {Uesato, Jonathan and Kushman, Nate and Kumar, Rohit and Song, Francis and Siegel, Noah and Wang, Luyu and Creswell, Antonia and Irving, Geoffrey and Higgins, Irina},
  journal = {arXiv preprint arXiv:2211.14275},
  year    = {2022},
  url     = {https://arxiv.org/abs/2211.14275}
}

@article{peng2025recap,
      title={Large Reasoning Models Learn Better Alignment from Flawed Thinking}, 
      author={ShengYun Peng and Eric Smith and Ivan Evtimov and Song Jiang and Pin-Yu Chen and Hongyuan Zhan and Haozhu Wang and Duen Horng Chau and Mahesh Pasupuleti and Jianfeng Chi},
      year={2025},
      eprint={2510.00938},
      archivePrefix={arXiv},
      primaryClass={cs.LG},
      url={https://arxiv.org/abs/2510.00938}, 
}

@inproceedings{lightman2024lets,
  title     = {Let's Verify Step by Step},
  author    = {Lightman, Hunter and Kosaraju, Vineet and Burda, Yura and Edwards, Harri and Baker, Bowen and Lee, Teddy and Leike, Jan and Schulman, John and Sutskever, Ilya and Cobbe, Karl},
  booktitle = {Proceedings of the International Conference on Learning Representations (ICLR)},
  year      = {2024},
  url       = {https://arxiv.org/abs/2305.20050}
}

@article{wang2024prm,
  title         = {Beyond the First Error: Process Reward Models for Reflective Mathematical Reasoning},
  author        = {Wang, Zhuoyan and others},
  journal       = {arXiv preprint},
  year          = {2025},
  eprint        = {2505.14391},
  archivePrefix = {arXiv},
  primaryClass  = {cs.CL},
  url           = {https://arxiv.org/abs/2505.14391}
}

@article{ssc_math2024,
  title         = {S$^3$c-Math: Spontaneous Step-level Self-correction Makes Large Language Models Better Mathematical Reasoners},
  author        = {Yan, Yuchen and Jiang, Jin and Liu, Yang and Cao, Yixin and Xu, Xin and Zhang, Mengdi and Cai, Xunliang and Shao, Jian},
  journal       = {arXiv preprint},
  year          = {2024},
  eprint        = {2409.01524},
  archivePrefix = {arXiv},
  primaryClass  = {cs.CL},
  url           = {https://arxiv.org/abs/2409.01524}
}

@article{lema2023,
  title  = {Learning From Mistakes Makes LLM Better Reasoner},
  author = {{LeMa Authors}},
  journal = {arXiv preprint},
  year   = {2023},
  url    = {https://www.semanticscholar.org/paper/98b607e7cb84e1a5c87c8a49562ae35435e6722d}
}

@article{xu2024pedcot,
  title         = {LLMs can Find Mathematical Reasoning Mistakes by Pedagogical Chain-of-Thought},
  author        = {Jiang, Zhuoxuan and Peng, Haoyuan and Feng, Shanshan and Li, Fan and Li, Dongsheng},
  journal       = {arXiv preprint},
  year          = {2024},
  eprint        = {2405.06705},
  archivePrefix = {arXiv},
  primaryClass  = {cs.CL},
  url           = {https://arxiv.org/abs/2405.06705}
}

@inproceedings{hu2025rmath,
  title     = {RMath: A Logic Reasoning-Focused Dataset Toward Mathematical Multistep Reasoning Tasks},
  author    = {Hu, Ziyi and Liu, Jun and Liu, Zhongzhi and Liu, Yuzhong and Xie, Zheng and Song, Yiping},
  booktitle = {Proceedings of the AAAI Conference on Artificial Intelligence},
  volume    = {39},
  number    = {22},
  pages     = {34585--34593},
  year      = {2025},
  doi       = {10.1609/aaai.v39i22.34585},
  url       = {https://doi.org/10.1609/aaai.v39i22.34585}
}

@inproceedings{you2024mumath,
  title     = {MuMath: Multi-perspective Data Augmentation for Mathematical Reasoning},
  author    = {You, Weilin and others},
  booktitle = {Findings of the Association for Computational Linguistics: NAACL 2024},
  year      = {2024},
  url       = {https://aclanthology.org/2024.findings-naacl.185}
}

@inproceedings{liu2025highmath,
  title     = {{H}igh{MATH}: Evaluating Math Reasoning of Large Language Models in Breadth and Depth},
  author    = {Liu, Yan and Zhang, Minghui and Xiong, Bojian and Xiao, Yifan and Sun, Yinong and Mei, Yating and Zeng, Longyu and Yang, Jingchao and Wang, Yang and Xiong, Deyi},
  booktitle = {Findings of the Association for Computational Linguistics: EMNLP 2025},
  year      = {2025},
  address   = {Suzhou, China},
  publisher = {Association for Computational Linguistics},
  url       = {https://aclanthology.org/2025.findings-emnlp.542}
}

@misc{xu2024symbcot,
      title={Faithful Logical Reasoning via Symbolic Chain-of-Thought}, 
      author={Jundong Xu and Hao Fei and Liangming Pan and Qian Liu and Mong-Li Lee and Wynne Hsu},
      year={2024},
      eprint={2405.18357},
      archivePrefix={arXiv},
      primaryClass={cs.CL},
      url={https://arxiv.org/abs/2405.18357}, 
}

@article{leang2024comat,
  title         = {CoMat: Chain of Mathematically Annotated Thought Improves Mathematical Reasoning},
  author        = {Leang, Joshua Ong Jun and others},
  journal       = {arXiv preprint},
  year          = {2024},
  eprint        = {2410.10336},
  archivePrefix = {arXiv},
  primaryClass  = {cs.CL},
  url           = {https://arxiv.org/abs/2410.10336}
}

@article{chen2022pot,
      title        = {Program of Thoughts Prompting: Disentangling Computation from Reasoning for Numerical Reasoning Tasks},
      author       = {Chen, Wenhu and Ma, Xueguang and Wang, Xinyi and Cohen, William W.},
      year         = {2023},
      eprint       = {2211.12588},
      archivePrefix= {arXiv},
      primaryClass = {cs.CL},
      url          = {https://arxiv.org/abs/2211.12588}
}

@article{panLogicLM23,
      title={Logic-LM: Empowering Large Language Models with Symbolic Solvers for Faithful Logical Reasoning}, 
      author={Liangming Pan and Alon Albalak and Xinyi Wang and William Yang Wang},
      year={2023},
      eprint={2305.12295},
      archivePrefix={arXiv},
      primaryClass={cs.CL},
      url={https://arxiv.org/abs/2305.12295}, 
}

@article{hendrycksmath2021,
      title        = {Measuring Mathematical Problem Solving With the MATH Dataset},
      author       = {Hendrycks, Dan and Burns, Collin and Kadavath, Saurav and Arora, Akul and Basart, Steven and Tang, Eric and Song, Dawn and Steinhardt, Jacob},
      year         = {2021},
      eprint       = {2103.03874},
      archivePrefix= {arXiv},
      primaryClass = {cs.LG},
      url          = {https://arxiv.org/abs/2103.03874}
}

@article{zhou2023leanstar,
      title={Lean-STaR: Learning to Interleave Thinking and Proving}, 
      author={Haohan Lin and Zhiqing Sun and Sean Welleck and Yiming Yang},
      year={2025},
      eprint={2407.10040},
      archivePrefix={arXiv},
      primaryClass={cs.AI},
      url={https://arxiv.org/abs/2407.10040}, 
}
\bibliographystyle{iclr2025_conference}










\end{document}